\documentclass{article}

\PassOptionsToPackage{numbers, compress}{natbib}

\usepackage[preprint]{neurips_2025}

\usepackage[utf8]{inputenc} 
\usepackage[T1]{fontenc}    
\usepackage{hyperref}       
\usepackage{url}            
\usepackage{booktabs}       
\usepackage{amsfonts}       
\usepackage{nicefrac}       
\usepackage{microtype}      
\usepackage{xcolor}         
\usepackage{amsmath}
\usepackage{graphicx}
\usepackage{enumitem} 
\usepackage{wrapfig}  
\usepackage{rotating}
\usepackage{float}
\usepackage{subcaption}
\title{Mitigating Context Bias in Domain Adaptation for Object Detection using Mask Pooling}

\author{%
  Hojun Son \quad Asma Almutairi \quad Arpan Kusari\\
  University of Michigan Transportation Research Institute\\
  University of Michigan\\
  Ann Arbor, MI-48103 \\
  \texttt{\{hojunson, asmaalm, kusari\}@umich.edu} \\
}

\begin{document}

\maketitle

\begin{abstract}
  Context bias refers to the association between the foreground objects and background during the object detection training process. Various methods have been proposed to minimize the context bias when applying the trained model to an unseen domain, known as domain adaptation for object detection (DAOD). But a principled approach to understand why the context bias occurs and how to remove it has been missing. 
  In this work, we provide a causal view of the context bias, pointing towards the pooling operation in the convolution network architecture as the possible source of this bias. We present an alternative, Mask Pooling, which uses an additional input of foreground masks, to separate the pooling process in the respective foreground and background regions and show that this process leads the trained model to detect objects in a more robust manner under different domains. We also provide a benchmark designed to create an ultimate test for DAOD, using foregrounds in the presence of absolute random backgrounds, to analyze the robustness of the intended trained models. Through these experiments, we hope to provide a principled approach for minimizing context bias under domain shift. 
\end{abstract}

\section{Introduction}
Domain adaptation for object detection (DAOD) addresses performance degradation caused by domain shifts between training and testing datasets. Despite advances in object detector architectures \cite{he2017mask, khanam2024yolov11, ren2016faster, lin2017focal, tan2019efficientnet, liu2016ssd} and training strategies \cite{luo2024unified, wang2025method, gevorgyan2022siou, shermaine2025image, zoph2020learning, triantafyllidou2024improving}, models still struggle to generalize to unseen domains \cite{weinstein2022general, kay2022caltech, weinstein2021benchmark}. Manual annotation on new datasets is expensive and labor-intensive. To alleviate this, sim-to-real approaches leverage synthetic data for training and real-world evaluation \cite{triess2022realism, horvath2022object, kainova2023overview}. However, the performance gap—known as the sim-to-real gap—remains substantial, and no method has surpassed the ``ORACLE" model trained on fully labeled target data \cite{kay2025align}. This is largely due to spurious factors and distributional perturbations learned during training \cite{wu2023discover, Xu_2023_CVPR}. 
DAOD algorithms mitigate domain shifts through feature alignment \cite{ganin2015unsupervised, chen2018domain, chen2021scale, zhu2019adapting, guan2021uncertainty}, adversarial domain classifiers \cite{ganin2016domain, Xu_2023_CVPR}, knowledge distillation \cite{kay2025align, pham2022revisiting, caron2021emerging, chen2017learning}, and hybrid methods \cite{cao2023contrastive, deng2021unbiased, xue2023cross, hoyer2023mic}. While these techniques aim to extract target-domain priors, these techniques still regenerate context bias between foreground (FG) and background (BG) during adaptation \cite{son2024quantifying}.

Causal tools can provide a natural way to measure the effect of the association between the FG and BG. In particular, Structural Causal Models (SCMs) provide a principled framework for capturing associational, interventional, and counterfactual causality \cite{scholkopf2021toward, kusner2017counterfactual, kilbertus2017avoiding, zhang2018fairness, karimi2020algorithmic, von2022fairness}, which traditional explainable AI (xAI) methods are not equipped to handle \cite{lundberg2017unified, chen2016xgboost, martin2021predicting, martin2021implicit}. Ultimately, modeling interventions to find causal discovery supports the development of predictive models that are robust to distributional shifts commonly encountered in real-world settings but also enables that can be extended to support reasoning.

Based on this knowledge, several recent studies have applied causal interventions to DAOD \cite{zhang2024causal, huang2021causal, resnick2021causal, jiang2023decoupled, lin2022causal, li2023disentangle, xu2023multi, chen2022c, li2022sigma, singh2020don}, with varying interpretations of context. One approach treats context as a causal factor of the image and performs intervention using joint representations and attention mechanisms \cite{huang2021causal}. Visual causality meaning frequent existence pairs in images and awareness of challenging group were modeled by causal intervention \cite{jiang2023decoupled}. Other study addresses contrast and spatial distribution biases as confounding factors in adaptation \cite{lin2022causal}, or identify and mitigate target discrimination bias in teacher-student frameworks using conditional causal intervention \cite{li2023disentangle}. Spurious correlations have also been targeted through frequency-domain augmentation and multi-view adversarial discriminators to isolate domain-invariant causal features \cite{xu2023multi}. Additionally, ambiguous boundaries are resolved by modeling a category-causality chain and intervening with saliency-based causal maps \cite{chen2022c}. The SIGMA framework \cite{li2022sigma} reformulates DAOD as a semantic-complete graph matching problem to address mismatched semantics and intra-class variance. While not a formal application of causal intervention (e.g., using do-calculus or SCMs), SIGMA adopts causal reasoning principles to reduce spurious correlations caused by missing semantics or noisy BGs.


However, it remains uncertain whether the performance gains are derived from the underlying causal hypotheses or simply due to increased model complexity, such as the addition of more parameters or layers. In particular, none of the previous research have taken a critical look at the pooling operation in Convolutional Neural Network (CNN) architectures which combines features for a particular layer without any prior discrimination of FG and BG features. While various efficient pooling methods have been proposed to better capture contextual information during training \cite{kobayashi2019gaussian, xu2022liftpool, zhai2017s3pool, zeiler2013stochastic,  liu2022graph, stergiou2022adapool, gao2019lip, stergiou2021refining}, these approaches often abstract low-level features into high-level representations that may inadvertently encode contextual biases as semantic cues. In particular, max pooling indiscriminately selects the highest activation within a kernel, regardless of its spatial origin, which can introduce FG-BG bias and degrade performance \cite{xu2022liftpool}.

These facts raise a critical question: \textit{can commonly used pooling operations in CNNs introduce artifacts that confound learning and contribute to domain shift?} Despite extensive research on alternative pooling techniques, their causal implications remain unexplored. We hypothesize that FG and BG features, being inherently independent, may form spurious associations during training, ultimately impairing generalization to novel domains.


To investigate this, we designed experiments leveraging object detection tasks to examine the causal relationship between FG-BG associations and model performance. We applied causal intervention-based analysis using ground-truth FG masks to isolate effects and draw causal inferences. We focused on multi objects detection because it provides a balanced complexity—more intricate than image classification, yet more manageable than semantic segmentation—making it suitable for causal analysis. Our findings offer insights for developing models with stronger domain generalization and out-of-distribution (OOD) robustness. We hope that our findings and experimental outcomes can provide remarkable perspectives on DAOD community. 

Our main contributions are as follows:
\begin{itemize}[noitemsep, topsep=0pt]
    \item We demonstrate that pooling operations can introduce causal artifacts that negatively affect model performance for DAOD task.
    \item We propose an alternate pooling strategy, Mask Pooling, to remove the causal effects of FG-BG associations on object detection outcomes and showcase the robustness empirically in different source and target domains using multiple neural network models. 
   \item We present a benchmark to evaluate model robustness for DAOD task by creating synthesized images with known FGs juxtaposed on random BGs. 
\end{itemize}
Through these contributions, we aim to provide a novel perspective to the DAOD community to mitigate context-bias using source-only domain. The paper is arranged as follows: Section \ref{sec:problem} defines the problem in causal terms, section \ref{sec:method} provides the pooling, models, datasets, evaluation and perturbation, section \ref{sec:experiment} provides the experimental results and section \ref{sec:discuss} provides the discussion. 

\section{Problem definition}
\label{sec:problem}
\begin{figure*}[t!]
\centering
\includegraphics[width=0.9\textwidth]{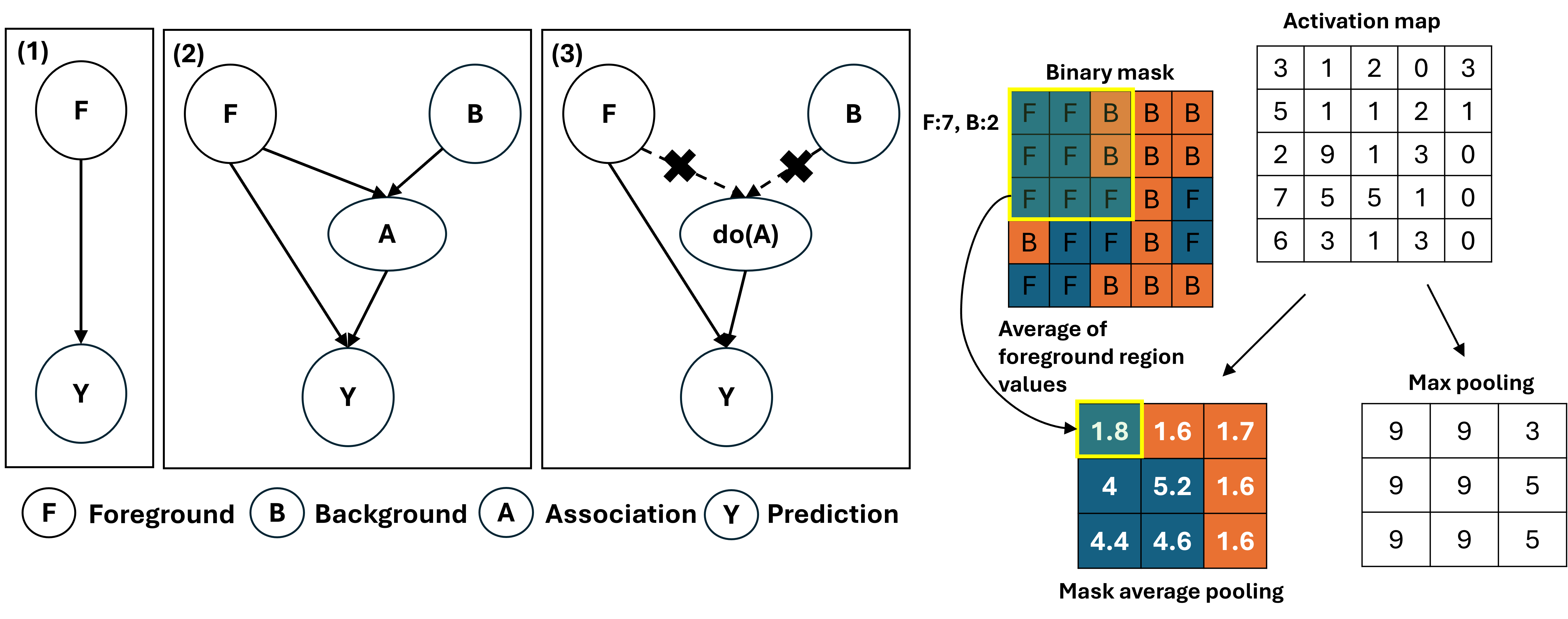}
\caption{\textbf{Left panel}: (1) shows the ideal case where the label ``Y" is dependent only on the FG (``F"); (2) presents the actual case where the FG (``F") and BG (``B") gets associated in ``A" and influences the prediction; (3) presents our proposed model where the mask pooling removes the association by making the pooling operation separately in FG and BG. \textbf{Right panel}: It shows an example activation map with max pooling and mask pooling to show the difference between the pooling techniques.}
\label{fig:DAG}
\end{figure*}

The left panel of Fig. \ref{fig:DAG} illustrates three causal cases in DAOD to investigate context bias. Case (1) (``F" $\rightarrow$ ``Y") represents cause-effect, where predictions rely solely on the distribution of FG features. Case (2) reflects findings from a quantification study in object detection \cite{son2024quantifying}, modeled as a v-structure: ``F" $\rightarrow$ ``A" $\leftarrow$ ``B"), where “A” encodes an implicit association between FG (“F”) and BG (“B”), often unobservable.  The prediction target ``Y" is influenced by both ``F" and ``A": ``F" $\rightarrow$ ``Y" $\leftarrow$ ``A". The corresponding joint distribution can be factorized as: $P(Y, A, F, B) = P(Y \mid F, A) \, P(A \mid F, B) \, P(F) \, P(B)$. Case (3) introduces causal intervention to analyze context bias by intervening on “A”. This yields the interventional distribution: $P(Y, F \mid do(A))) = P(Y \mid F, do(A)))P(F)$ which simplifies to: $P(Y, F) \approx P(Y \mid F) \cdot P(F) \approx P(F \mid Y) \cdot P(Y)$. With respect to Maximum likelihood, by the intervention, a model will be trained increase the likelihood based on FG features because it follows anti-causal process verified mathematically  \cite{kugelgen2020semi, janzing2015semi}. Besides, the study in \cite{son2024quantifying} demonstrated that FG-to-FG features across domains are more similar than BG-related combinations, enabling robustness even target domains are unseen during training.

To investigate this, we compare models trained with and without FG masks. The use of masks enforces determinism in the variable ``A" by explicitly separating FG from BG, thereby constraining the influence of spurious contextual features. Through causal intervention, we analyze the effectiveness of mitigating context bias and assess the robustness achieved by FG and BG separation.

\section{Methodology}
\label{sec:method}
The preceding sections have revealed hypothesis and the literature gap between DAOD and causality analysis in CNN-based deep learning algorithms. Based on the hypothesis, we designed experiments to evaluate and quantify.

\subsection{Boundary-aware pooling}
\label{sec:boundary_aware}
\begin{equation}
\renewcommand{\arraystretch}{2}
P_{m,n} = \left\{
\begin{array}{ll}
    \displaystyle\frac{1}{n_F} \sum\limits_{(i,j) \in K} F(i,j) \cdot x(i,j) & \quad \text{if } n_F \geq n_B \\[12pt]
    \displaystyle\frac{1}{n_B} \sum\limits_{(i,j) \in K} (1 - F(i,j)) \cdot x(i,j) & \quad \text{otherwise}
\end{array}
\right.
\label{eq:boundary_pooling}
\end{equation}

A FG mask is a binary mask of all FG objects. For example, Cityscapes \cite{cordts2016cityscapes} has 8 different FG classes. Given the masks, the pooling operates to compute average and voting based on the masks. Where \( P_{m,n} \) is the pooled value, \( K \) is the \( 3 \times 3 \) kernel region, \( x(i,j) \) is the pixel value at position \( (i,j) \), and \( F(i,j) \in \{0,1\} \) indicates whether \( (i,j) \) is a FG pixel (1) or a BG pixel (0). $n_F$ is the number of FG pixels and $n_B$ is the number of BG pixels in the mask respectively. The designed pooling separates each region using ground truth masks into FG and BG during training and inference and average of each region is transferred respectively depending on voting systems. Fig. \ref{fig:DAG} and equation \ref{eq:boundary_pooling} illustrate how our proposed pooling layer works.

\subsection{Neural network models}
We compare two backbone architectures: ResNet-50 \cite{he2016deep} and EfficientNet-B0 \cite{tan2019efficientnet}. ResNet-50 includes a single max pooling layer after the stem, whereas EfficientNet-B0 does not contain any explicit pooling layer except global average pooling. In our setup, the max pooling in ResNet-50 is replaced with a mask pooling operation inserted after the first block, creating the modified version of ResNet-50, which we refer to as ``ResM". For EfficientNet-B0, mask pooling is added after the stem layer - creating ``EffM", which also results in a reduction in GFLOPs. Both backbones are integrated into the Detectron2 \cite{Detectron2018} framework using a Feature Pyramid Network (FPN)-based architecture \cite{lin2017feature}. ResNet-50 is initialized with COCO-pretrained weights, while EfficientNet-B0 uses ImageNet-1K pretrained weights. We compare performance against the state-of-the-art DAOD model, ALDI++ (Backbone: ResNet-50 FPN) \cite{kay2025align}. In total, five model variants are evaluated. Note that global average pooling layers in EfficientNet remain unchanged.

\subsection{Datasets}
We used the following datasets in this paper:
\begin{itemize}[noitemsep, topsep=0pt]
\item Cityscapes \cite{cordts2016cityscapes}: Large-scale urban dataset with 2975 training images with 500 validation images and 1525 finely annotated test images. There are 8 FG object classes and 11 BG object classes. We used 500 validation for our experiments. Cityscapes foggy \cite{SDV18} and Cityscapes rainy \cite{hu2019depth} dataset are synthetically augmented on Cityscapes. We used 1500 foggy validation and evaluated on 1188 rainy validation images for our experiment.
\item KITTI Semantic Segmentation \cite{Alhaija2018IJCV}: Large-scale driving dataset collected in Karlsruhe, Germany. The semantic segmentation subset contains 200 finely annotated images for training with pixel-level semantic labels. Similar to Cityscapes, there are 8 FG and 11 BG classes. We used KITTI Semantic train only for our evaluation. 
\item Virtual KITTI \cite{gaidon2016virtual}: A photo-realistic synthetic images based on KITTI Object tracking scenes. It consists of 6 different weather conditions and 4 different translation augmentation. We used only weather conditions dataset with 2126 images and 375x1242 image resolution. It has 3 FG labels and 10 BG classes.
\item BG-20K \cite{li2022bridging}: A dataset with 20,000 high-resolution BG images. We utilize this dataset to create randomly chosen BG images combined with FG objects in Cityscapes, Cityscapes foggy, Cityscapes rainy, KITTI Semantic, and Virtual KITTI related dataset. 
\end{itemize}
Here are the abbreviations for the datasets we defined in this paper:
\vspace{-0.25in}
\begin{table}[!h]
    \centering
    \caption{Dataset abbreviations}
    \begin{tabular}{c c}
    \hline
    Abbreviation & Meaning \\
    \hline
      CV/ CFV / CRV   &  Cityscapes Validation / Foggy / Rainy \\
      KST  &  KITTI Semantic Train \\
      BG-20K  &   Background 20K Dataset \\
      ALDI++ & Resnet-50 FPN with ALDI++ \\
      Res / ResM  &  Resnet-50 FPN / with mask pooling \\
      Eff / EffM  &  EfficientNet-B0 FPN / with mask pooling \\
       VKC / VKF / VKM /   &   Virtual KITTI Clone / Foggy / Morning / \\
       VKO / VKR / VKS & Virtual KITTI Overcast / Rain / Sunset \\
       \hline
    \end{tabular}
    \label{tab:abb}
\end{table}

\subsection{Causal analysis}
In order to infer the FG-BG association by causal intervention, we design three separate experiments to empirically validate our hypothesis:


\subsubsection{BG activation map perturbation}
We modified activation maps during inference and multiplied small weight to simulate BG perturbation in feature space. It demonstrates robustness of trained models and on context bias between FG and BG. These studies \cite{ramaswamy2020ablation, meng2021foreground} suggested FG activation is related to performance. We adjusted activation values using BG masks from 0.5 to 2.75 with 0.25 step size.

\subsubsection{Random BG assignment with FG}
While there could be associations between common objects such as ``car" and ``road", we are interested in evaluating if the car is detected in absence of road. Therefore, we design a test to sample random BGs for each image frame and then placing the FGs into the BG image using blending operation. The idea is then to check how many of these FGs are detected given these modified images. Large scale inference of such random placement can give us a measure of the robustness of the models. We evaluated on total 20 different datasets (3 Cityscapes related, ``KST", and 6 Virtual KITTI related and with and without BG-20K compound images. The synthetic datasets consists of randomly chosen BG images compounded with FG images using FG masks.
All images have different BG images.

\subsubsection{Fixed random BG assignment with FG}
While the previous experiment provides a distributional measure of the robustness, it does not provide a measure of the performance of a model for different FGs. Therefore, in this experiment, we utilize a single randomly chosen BG image for the 20 datasets and provide results on them (5 repetition with 5 sampled BG images). This helps us provide a deterministic BG information on which to understand the performance of the model for different FGs. 

\section{Experiment}
\label{sec:experiment}
In this section, we provide the implementation details including the evaluation metrics, number of evaluations and training parameters for each model. 
\subsection{Evaluation metric}
We utilize two different evaluation metrics:
\begin{itemize}
    \item mAP50 - Mean Average Precision at a threshold 50 is standard for detection tasks \cite{kay2025align}. We logged the mAP50 of each model for the different experiments to understand the trends.  
    \item Hierarchical F1 score \cite{riehl2023hierarchical} -  In order to understand the effect of changing the BG activations or inputs, we need to provide a measure of change due to the intervention (how many are detected after the intervention as opposed to before the intervention). Therefore, we select the hierarchical F1 score which provides a classification score for hierarchical problems.
\end{itemize}

\subsection{Experimental procedure}
\begin{enumerate}
    \item Evaluated 5 models including ALDI++ given by official repository on 20 different validation set. Only ``VKC" is a training set and validation set.
    To avoid randomness effect, we measured 5 times evaluation on the synthetic dataset for various BG. 
    \item Computed hierarchal F1 score variations on Cityscapees trained set and ``VKC".
    \item Measured changes of mAP50 depending on variations of BG activation perturbation.
\end{enumerate}
\textbf{Training parameters}: We utilized the Detectron2 framework to train the models described above on the Cityscapes training set, other than ALDI++ which comes pretrained on Cityscapes. For Virtual KITTI datasets, we trained our models with the same training parameters while ALDI++ was trained with ``VKC" as source domain and ``VKF" as target domain using the best training strategy.
We utilized a learning rate of 0.02 for ``Res" and ``ResM" and 0.01 for ``Eff" and ``EffM", with a max iteration of 10000 for ``Res" and ``ResM" and 20000 for ``Eff" and ``EffM". For ALDI++, all training parameters were maintained except max iteration. We applied the different data augmentation techniques for burn-in in ALDI++ except cutout \cite{devries2017improved}. 
We used a batch size of 8 with an image resolution of 1024x2048. We trained and evaluated the model on NVIDIA RTX A4500 20GB without any change in loss function.

\subsection{AP50 measurement results}
\label{sec:ap50}
We remeasured released ALDI++ mAP50 trained on Cityscapes training set, for ``CRV" and ``KST". ALDI++ outperformed other models on``CFV" and ``CRV". It has a benefit of using target domain information obtained from Cityscapes foggy training set to learn foggy features. ``ResM" superior to other models on ``CV" and ``KST". Notably, it demonstrated superior outcomes on ``KST" compared ALDI++. Even ``EffM" outperformed ALDI++ by about 2\%  with lowest computational costs among other models (see Table \ref{tab:cityscapes_AP_50_mean_std_results_wo_bg_20k}). With Cityscapes + ``BG" datasets, ``ResM" is superior than other models on all datasets and ``EffM" demonstrated significant difference from ALDI++ on ``KST+BG" about 5.5\%. (see Table \ref{tab:cityscapes_AP_50_mean_std_results_w_bg_20k}). Moreover, ``EffM" initialized on the Imagenet1K pretrained model \cite{deng2009imagenet} which is less sustainable than COCOdataset \cite{lin2014microsoft} pretrained model. 1024x2048 image resolution was used to measure GFLOPs and the number of parameters (``Params"). ``Res" has 398.5 GFlops and 41.3M parameters, ALDI++ and ``ResM"" have the same GFLOPs and parameters with ``Res". ``Eff" has 237.7 GFLOPs and 20.6M parameters and ``EffM" shows 236.7 GFLOPs and 20.6M parameters. 

\begin{figure*}[t!]
\centering
\includegraphics[width=0.8\textwidth]{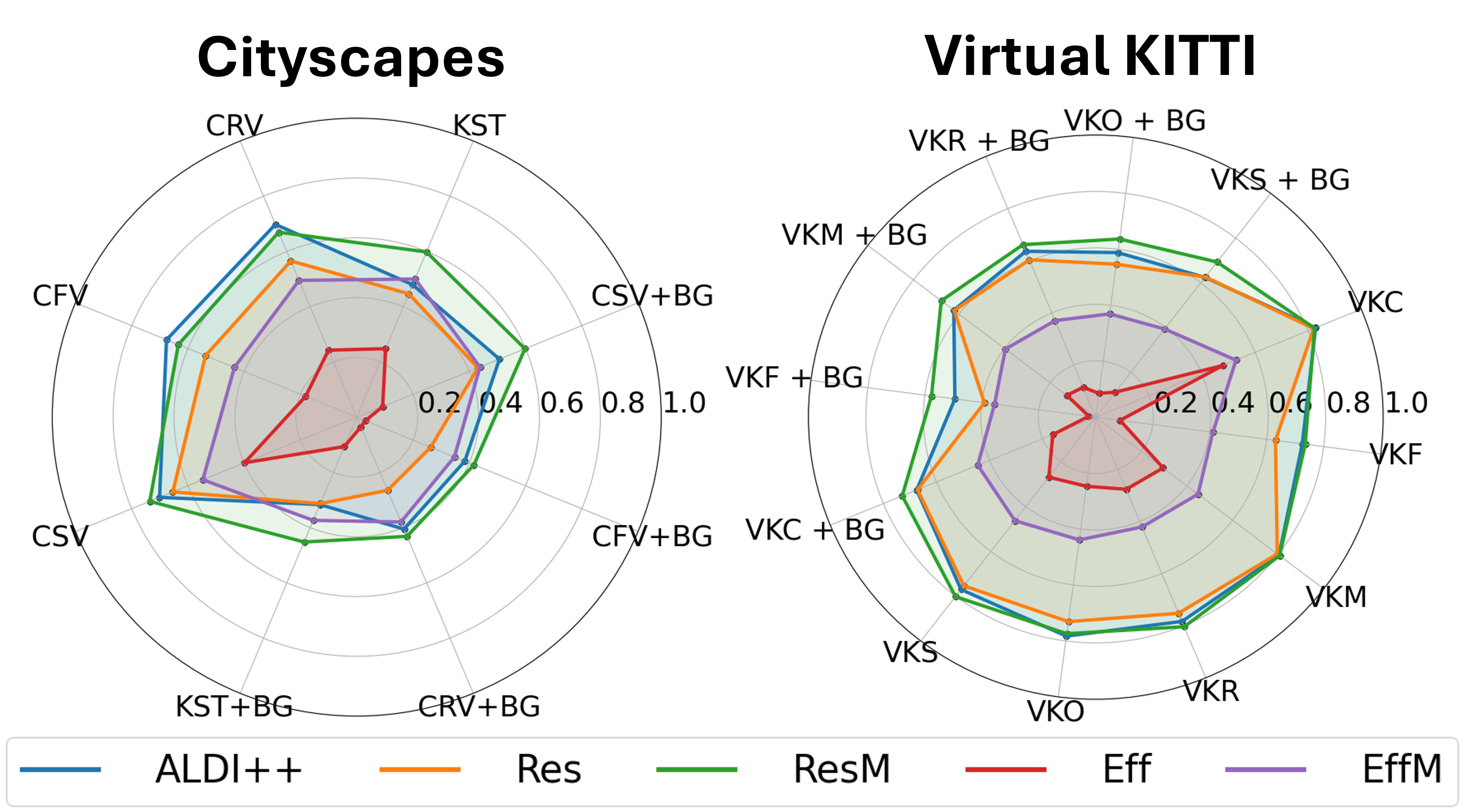}
\caption{\textbf{Radar chart of mAP50 on Cityscapes and Virtual KITTI}. It is normalized by 100.}
\label{fig:radar_chart_performance_ap50}
\end{figure*}

With Virtual KITTI datasets, even though ALDI++ has a benefit of using target domain information (``VKF"), it could not beat ``ResM" on ``VKF" (see Table \ref{tab:virtual_kitti_ap50_results}).  ``ResM" demonstrated its robustness across all synthetic dataset (see Table \ref{tab:virtual_kitti_bg_results}). ``EffM" could not outperformed ALDI++ in this case. Fig. \ref{fig:radar_chart_performance_ap50} illustrates 20 different mAP50 in two radar charts. 

Table \ref{tab:fixed_bg_mAP50} demonstrates mAP50 with all dataset compounded on single randomly chosen images from ``BG-20K". ``ResM" is superior to other models. 

\begin{table*}[t!]
\centering
\caption{\textbf{mAP50 scores on various Cityscapes dataset.} "reported" means the value from the published paper.}
\begin{tabular}{cccccc} \hline  

\textbf{Dataset} & \textbf{ALDI++} & \textbf{Res} & \textbf{ResM} & \textbf{Eff} & \textbf{EffM}  \\ \hline  

CV & 70.093 (reported:NA) & 65.386 & \textbf{73.572} & 39.884 & 54.723  \\  
CFV & \textbf{67.626} (reported: 66.8) & 53.770 & 63.413 & 18.174 & 43.492  \\  
CRV & \textbf{69.790} (reported: NA) & 56.555 & 66.971 & 24.255 & 49.524  \\  
KST & 47.965 (reported: NA) & 44.674 & \textbf{59.821} & 24.798 & 50.027  \\
\hline
\end{tabular}
\label{tab:cityscapes_AP_50_mean_std_results_wo_bg_20k}
\end{table*}

\begin{table*}[t!]
\centering
\caption{\textbf{Mean ± standard deviation of mAP50 across synthetic datasets and models.}}
\begin{tabular}{cccccc}
\hline
\textbf{Dataset} & \textbf{ALDI++} & \textbf{Res} & \textbf{ResM} & \textbf{Eff} & \textbf{EffM} \\
\hline
CV + BG & 50.74 ± 1.02 & 43.10 ± 1.09 & \textbf{59.80} ± 1.10 & 9.21 ± 0.32 & 43.80 ± 0.35 \\
CFV + BG & 38.42 ± 0.40 & 26.29 ± 0.21 & \textbf{41.67} ± 0.07 & 3.09 ± 0.17 & 34.73 ± 0.29 \\
CRV + BG & 40.52 ± 0.27 & 26.49 ± 0.45 & \textbf{43.16} ± 0.20 & 3.53 ± 0.19 & 37.97 ± 0.43 \\
KST + BG & 31.66 ± 1.09 & 31.29 ± 0.93 & \textbf{45.20} ± 0.38 & 10.66 ± 0.91 & 37.37 ± 1.30 \\
\hline
\end{tabular}
\label{tab:cityscapes_AP_50_mean_std_results_w_bg_20k}
\end{table*}

\subsection{Hierarchical F1 score results}
\label{sec:hierarchical_f1}
Computed average hierarchical F1 score across with BG modification indicate performance gains from removal of FG-BG association by comparing each model trained with mask pooling and without mask pooling. Interesting finding is bicycle and motorcycle have new association because $Diff$ is negative in ``Res" and ``ResM" comparison. The average hierarchical F1 score of all FG object in comparison between ``Eff" and ``EffM" is significantly improved.  Table \ref{tab:f1_score} and \ref{tab:merged_fg_metrics} demonstrate the F1 score differences per FG object. "ResM" achieved F1 score improvements in 64 out of 88 evaluation pairs, while "EffM" improved in 76 out of 77 pairs, primarily due to its failure to detect the train class in ``Eff". On the "VKC" dataset, "ResM" showed 19 improvements out of 30 cases, whereas "EffM" achieved 29. The F1 score difference ($\mathit{Diff}$) between "Res" and "ResM" is marginal, while the difference between "Eff" and "EffM" is substantial.



\begin{table}[ht!]
\setlength{\tabcolsep}{7pt}
\centering
\caption{\textbf{Hierarchical F1 score improvements across different backbones on Cityscapes train set}. Dashes denote missing values.}
\begin{tabular}{ccccccc}
\hline
\textbf{fg} & \textbf{Res} & \textbf{ResM} & \textbf{Diff} & \textbf{Eff} & \textbf{EffM} & \textbf{Diff} \\
\hline
bicycle     & 0.764836 & 0.703582 & -0.061255 & 0.392509 & 0.542873 & 0.150364 \\
bus         & 0.731709 & 0.769673 &  0.037964 & 0.076018 & 0.446318 & 0.370300 \\
car         & 0.846873 & 0.879145 &  0.032273 & 0.830591 & 0.835645 & 0.005055 \\
motorcycle  & 0.655818 & 0.601345 & -0.054473 & 0.398436 & 0.662518 & 0.264082 \\
person      & 0.703664 & 0.818418 &  0.114755 & 0.433718 & 0.698645 & 0.264927 \\
rider       & 0.692245 & 0.700255 &  0.008009 & 0.579636 & 0.721009 & 0.141373 \\
train       & 0.361891 & 0.853300 &  0.491409 & -        & -        & -        \\
truck       & 0.647836 & 0.665127 &  0.017291 & 0.021545 & 0.398591 & 0.377045 \\
\hline
\end{tabular}
\label{tab:f1_score}
\end{table}

\begin{table}[t!]
\setlength{\tabcolsep}{7pt}
\centering
\caption{\textbf{Hierarchical F1 score improvements across different backbones on ``VKC"}.}
\begin{tabular}{ccccccc}
\toprule
\textbf{fg} & \textbf{Res} & \textbf{ResM} & \textbf{Diff} & \textbf{Eff} & \textbf{EffM} & \textbf{Diff} \\
\midrule
Car   & 0.757500 & 0.755610 & –0.001890 & 0.591700 & 0.606220 & 0.014520 \\
Truck & 0.962700 & 0.952020 & –0.010680 & 0.763660 & 1.136580 & 0.372920 \\
Van   & 0.834670 & 0.828100 & –0.006570 & 0.594020 & 0.737010 & 0.142990 \\
\bottomrule
\end{tabular}
\label{tab:merged_fg_metrics}
\end{table}

\begin{table}[t!]
\centering
\caption{\textbf{mAP50 scores on various Virtual KITTI.}}
\begin{tabular}{lccccc}
\toprule
Dataset & \textbf{ALDI++} & \textbf{Res} & \textbf{ResM} & \textbf{Eff
}& \textbf{EffM}  \\
\midrule
VKC    & \textbf{82.797} & 82.075 & 82.412 & 47.875 & 52.811  \\
VKF     & 72.492 & 63.059 & \textbf{73.512} &  8.529 & 41.095  \\
VKM  & 80.369 & 79.626 & \textbf{80.630} & 29.549 & 44.846  \\
VKR  & 78.357 & 75.238 & \textbf{80.392} & 27.669 & 42.116  \\
VKO & \textbf{78.285} & 73.098 & 77.409 & 24.721 & 43.868  \\
VKS  & 77.003 & 75.357 & \textbf{80.177}& 26.815 & 46.240  \\
\bottomrule
\end{tabular}
\label{tab:virtual_kitti_ap50_results}
\end{table}
\vspace{-10pt}

\begin{table}[t!]
\centering
\caption{\textbf{Mean ± standard deviation of mAP50 across synthetic datasets and models.}}
\begin{tabular}{lccccc}
\toprule
\textbf{Dataset} & \textbf{ALDI++} & \textbf{Res} & \textbf{ResM} & \textbf{Eff} & \textbf{EffM}  \\
\midrule
VKC + BG & $67.45 \pm 0.18$ & $67.14 \pm 0.16$ & \textbf{72.90} $\pm 0.35$ & $16.19 \pm 0.18$ & $44.46 \pm 0.29$  \\
VKF + BG & $49.44 \pm 0.21$ & $39.11 \pm 0.56$ & \textbf{57.66} $\pm 0.28$ &  $2.45 \pm 0.11$ & $35.52 \pm 0.11$  \\
VKM + BG & $62.28 \pm 0.25$ & $62.08 \pm 0.34$ & \textbf{67.77} $\pm 0.67$ & $12.69 \pm 0.47$ & $39.64 \pm 0.45$  \\
VKR + BG & $63.64 \pm 2.10$ & $60.36 \pm 0.42$ & \textbf{66.19} $\pm 0.40$ & $11.48 \pm 0.13$ & $37.04 \pm 0.30$  \\
VKO + BG & $58.83 \pm 0.49$ & $54.69 \pm 0.71$ & \textbf{63.69} $\pm 0.54$ &  $8.50 \pm 0.11$ & $36.92 \pm 0.24$  \\
VKS + BG & $62.45 \pm 0.21$ & $62.62 \pm 0.24$ & \textbf{69.32} $\pm 0.23$ & $11.16 \pm 0.16$ & $39.38 \pm 0.31$  \\
\bottomrule
\end{tabular}
\label{tab:virtual_kitti_bg_results}
\end{table}
\vspace{-10pt}

\subsection{BG activation map perturbation results}
\label{sec:background_perturbation}
``ResM" and ALDI++ demonstrated its robustness through the experiment (see Table \ref{tab:background_perturbation}). ALDI++ showed the robustness on ``CFV" and ``CRV" because it has target domain information (Cityscapes Foggy train). However, ``ResM" demonstrated comparable diff with respect to ``CFV". It outperformed ALDI++ on``CV" and ``KST" which ALDI++ have never seen. On Vritual KITTI, ALDI++ was superior to all models but ``ResM" was dominant on Virtual KITTI + ``BG" dataset except on ``VKC + BG" with the tiny diff.


\begin{table}[t]
\centering
\caption{\textbf{mAP50 results with a fixed BG randomly selected from BG-20K.} Italic values mean lower Diff of ``EffM" than ALDI++.}
\footnotesize
\setlength{\tabcolsep}{8pt} 
\renewcommand{\arraystretch}{1.1}
\begin{tabular}{lccccc}
\toprule
\textbf{Dataset} & \textbf{ALDI++} & \textbf{Res} & \textbf{ResM} & \textbf{Eff} & \textbf{EffM} \\
\midrule
CV + 1 BG & 50.79 $\pm$ 10.13 & 47.63 $\pm$ 12.66 & \textbf{60.31} $\pm$ 7.82 & 8.16 $\pm$ 11.30 & 44.48 $\pm$ 1.29 \\
CFV+ 1 BG & 34.91 $\pm$ 14.09 & 31.26 $\pm$ 9.86 & \textbf{40.76} $\pm$ 6.06 & 5.59 $\pm$ 5.50 & 33.33 $\pm$ 6.77 \\
CRV+ 1 BG & \textbf{48.86} $\pm$ 7.78 & 28.91 $\pm$ 8.60 & 44.11 $\pm$ 8.59 & 6.02 $\pm$ 7.78 & 39.17 $\pm$ 8.75 \\
KST+ 1 BG & 34.96 $\pm$ 6.09 & 34.43 $\pm$ 8.18 & \textbf{44.36} $\pm$ 5.71 & 8.51 $\pm$ 10.61 & \textit{40.43} $\pm$ 1.38 \\
VKC+ 1 BG & 67.24 $\pm$ 4.32 & 68.27 $\pm$ 4.34 & \textbf{74.72} $\pm$ 0.99 & 16.19 $\pm$ 6.99 & 45.44 $\pm$ 0.86 \\
VKF+ 1 BG & 55.76 $\pm$ 7.85 & 39.39 $\pm$ 6.90 & \textbf{61.35} $\pm$ 2.34 & 3.13 $\pm$ 3.37 & 36.75 $\pm$ 2.11 \\
VKM+ 1 BG & \textbf{67.51} $\pm$ 4.76 & 64.86 $\pm$ 3.86 & 67.32 $\pm$ 3.77 & 14.68 $\pm$ 4.98 & 40.55 $\pm$ 1.73 \\
 VKO+ 1 BG & 65.98 $\pm$ 2.61 & 66.64 $\pm$ 2.81 & \textbf{69.02} $\pm$ 1.66 & 15.21 $\pm$ 7.07 & 39.28 $\pm$ 3.72 \\
VKR+ 1 BG & 60.00 $\pm$ 7.75 & 57.33 $\pm$ 5.61 & \textbf{67.14} $\pm$ 2.74 & 9.87 $\pm$ 5.71 & 37.86 $\pm$ 1.42 \\
VKS+ 1 BG & 65.19 $\pm$ 8.17 & 65.91 $\pm$ 2.76 & \textbf{71.76} $\pm$ 2.54 & 12.78 $\pm$ 4.74 & 42.86 $\pm$ 3.46 \\

\bottomrule
\end{tabular}
\label{tab:fixed_bg_mAP50}
\end{table}

\begin{table*}[t]
\centering
\caption{ \textbf{mAP50 range comparison (Min / Max / Diff) across 20 datasets with and without BG intervention.} Bold values indicate the lowest variation (Diff) in each dataset group. Lower Diff indicates robustness on BG perturbation. 
Italic values mean lower Diff of ``EffM" than ALDI++. The lowest mAP50 of ``EffM" (16.25) is disregarded due to lower Max mAP50.}
\setlength{\tabcolsep}{7pt}
\scriptsize
\begin{tabular}{lcccccccccccc}
\toprule
\textbf{Model} 
& \multicolumn{3}{c}{\textbf{CV}} & \multicolumn{3}{c}{\textbf{CFV}} & \multicolumn{3}{c}{\textbf{CRV}} & \multicolumn{3}{c}{\textbf{KST}} \\
& Min & Max & Diff & Min & Max & Diff & Min & Max & Diff & Min & Max & Diff \\
\midrule
ALDI++ & 63.07 & 71.82 & 8.75 & 59.05 & 67.64 & \textbf{8.60} & 67.13 & 71.11 & \textbf{3.98} & 25.87 & 50.32 & 24.45 \\
EffM   & 10.46 & 54.73 & 44.27 & 13.10 & 43.39 & 30.29 & 4.93  & 50.50 & 45.57 & 24.68 & 52.72 & 28.04 \\
ResM   & 66.27 & 73.57 & \textbf{7.30} & 54.75 & 63.60 & 8.85 & 60.12 & 66.97 & 6.85 & 43.32 & 60.19 & \textbf{16.86} \\
\bottomrule
\end{tabular}
\begin{tabular}{lcccccccccccc}
\textbf{Model} 
& \multicolumn{3}{c}{\textbf{CV+BG}} & \multicolumn{3}{c}{\textbf{CFV+BG}} & \multicolumn{3}{c}{\textbf{CRV+BG}} & \multicolumn{3}{c}{\textbf{KST+BG}} \\
& Min & Max & Diff & Min & Max & Diff & Min & Max & Diff & Min & Max & Diff \\
\midrule
ALDI++ & 30.74 & 58.75 & 28.01 & 22.95 & 45.29 & 22.34 & 27.87 & 46.44 &\textbf{18.58} & 15.66 & 38.30 & 22.63 \\
EffM   & 22.16 & 44.45 & \textit{22.28} & 18.19 & 34.44 & \textit{16.25} & 22.49 & 39.83 & \textit{17.34} & 20.58 & 41.03 & \textit{20.45} \\
ResM   & 42.06 & 62.23 & \textbf{20.17} & 26.60 & 45.19 & \textbf{18.59} & 26.81 & 46.54 & 19.73 & 28.23 & 45.51 & \textbf{17.28} \\
\bottomrule
\end{tabular}

\begin{tabular}{lcccccccccccc}
\textbf{Model} 
& \multicolumn{3}{c}{\textbf{VKC}} & \multicolumn{3}{c}{\textbf{VKF}} & \multicolumn{3}{c}{\textbf{VKM}} & \multicolumn{3}{c}{\textbf{VKO}} \\
& Min & Max & Diff & Min & Max & Diff & Min & Max & Diff & Min & Max & Diff \\
\midrule
ALDI++ & 76.69 & 82.30 & \textbf{5.61} & 57.30 & 72.49 & \textbf{15.19} & 71.53 & 80.37 & \textbf{8.84} & 71.87 & 78.36 & \textbf{6.49} \\
EffM   & 33.05 & 52.81 & 19.76 & 23.28 & 41.10 & 17.82 & 30.35 & 45.43 & 15.08 & 29.77 & 43.05 & 13.28 \\
ResM   & 72.81 & 82.41 & 9.61 & 43.87 & 73.51 & 29.64 & 69.87 & 80.63 & 10.76 & 66.62 & 80.39 & 13.78 \\
\bottomrule
\end{tabular}

\begin{tabular}{lcccccccccccc}
\textbf{Model} 
& \multicolumn{3}{c}{\textbf{VKR}} & \multicolumn{3}{c}{\textbf{VKS}} & \multicolumn{3}{c}{\textbf{VKC+BG}} & \multicolumn{3}{c}{\textbf{VKF+BG}} \\
& Min & Max & Diff & Min & Max & Diff & Min & Max & Diff & Min & Max & Diff \\
\midrule
ALDI++ & 64.63 & 78.29 & \textbf{13.65} & 66.90 & 77.00 & \textbf{10.10} & 49.25 & 70.61 & \textbf{21.36} & 25.40 & 55.86 & 30.46 \\
EffM   & 32.17 & 43.87 & 11.70 & 29.93 & 46.24 & 16.31 & 17.28 & 44.47 & 27.19 & 11.26 & 35.50 & \textit{24.24} \\
ResM   & 54.54 & 77.41 & 22.87 & 69.23 & 80.18 & 10.95 & 51.13 & 72.83 & 21.70 & 31.40 & 58.72 & \textbf{27.32} \\
\bottomrule
\end{tabular}

\begin{tabular}{lcccccccccccc}
\textbf{Model} 
& \multicolumn{3}{c}{\textbf{VKM+BG}} & \multicolumn{3}{c}{\textbf{VKO+BG}} & \multicolumn{3}{c}{\textbf{VKR+BG}} & \multicolumn{3}{c}{\textbf{VKS+BG}} \\
& Min & Max & Diff & Min & Max & Diff & Min & Max & Diff & Min & Max & Diff \\
\midrule
ALDI++ & 39.75 & 66.86 & 27.11 & 41.54 & 66.09 & 24.55 & 35.93 & 63.80 & 27.87 & 39.86 & 66.84 & 26.98 \\
EffM   & 16.48 & 39.62 & \textit{23.14} & 14.12 & 37.03 & \textit{22.91} & 13.17 & 36.90 & \textit{23.73} & 15.05 & 40.02 & \textit{24.97} \\
ResM   & 45.14 & 68.11 & \textbf{22.97} & 40.48 & 65.88 & \textbf{25.40} & 37.84 & 64.12 & \textbf{26.28} & 45.90 & 69.29 & \textbf{23.39} \\
\bottomrule
\end{tabular}

\label{tab:background_perturbation}
\end{table*}
\vspace{-10pt}


\subsection{Ablation study} 
We evaluated the impact of boundary perturbations using morphological operations on ``CV". Applying a dilation with a factor of 1.2, which introduces over 20\% boundary error, resulted in a performance drop from 73.578 to 53.800. A smaller dilation factor of 1.1, corresponding to approximately 10\% boundary error, reduced the score to 65.642. Similarly, applying erosion with a factor of 0.9 (about 10\% boundary error) decreased the performance to 64.002, while a more aggressive erosion factor of 0.8 (around 20\% error) further reduced it to 49.015. These results demonstrate the models learned boundary of FG features and predicted based on FG features.




\subsection{Benchmark for DAOD}
Our mAP50 tables serve as a comprehensive benchmark for evaluating DAOD performance. ALDI++ achieved state-of-the-art mAP50 on the "CFV" dataset; however, it was not evaluated on "CRV" or "KST". Moreover, ALDI++ did not report results on the Virtual KITTI or the synthetic datasets constructed with random BGs. Notably, Virtual KITTI differs from standard object detection datasets, as it is originally designed for object tracking and exhibits strong temporal correlations and spatial correlation across dataset. This characteristic introduces a distinct evaluation perspective for DAOD.

\section{Discussion and Conclusions}
\label{sec:discuss}
This work introduces a causally grounded analysis of context bias in DAOD, highlighting the pooling operation as a previously underexamined source of spurious FG-BG associations. We propose Mask Pooling as a principled intervention to explicitly separate FG and BG regions during feature aggregation, leading to improved generalization under domain shifts.

Our benchmark evaluations demonstrate that Mask Pooling consistently improves robustness under domain shifts, especially in synthetic setups where BGs are replaced with random samples (e.g., BG-20K). In particular, ``EffM" showed substantial improvements in hierarchical F1 scores over its baseline, achieving gains in 76 out of 77 evaluation pairs (see Section \ref{sec:hierarchical_f1}). These improvements are more pronounced than those observed between ``Res" and ``ResM", likely due to EfficientNet’s architectural characteristics: it lacks early-stage max pooling and uses depthwise convolutions, resulting in larger effective receptive fields.  This may amplify the impact of mask pooling, especially for fine-grained boundary preservation.

Mask Pooling offers critical advantages over conventional max pooling, which often amplifies the strongest activation without regard to semantic structure. As discussed in Section \ref{sec:boundary_aware} and evidenced in boundary perturbation experiments (see Section \ref{sec:background_perturbation}), max pooling can entangle FG and BG signals, whereas mask pooling explicitly constrains pooling to semantically meaningful regions.  Pooling is a feature extraction operation and thus, the intermingling of features happen during the extraction process itself. Therefore, we cannot use a masking operation after the feature extraction process. This results in better generalization, as shown by mAP50 improvements on Cityscapes and synthetic domains. Averaging over the FG region allows the model to extract a broader range of semantic features, in contrast to max pooling, which selects only the most distinctive activation and can introduce spike during training.

Our evaluation on the Virtual KITTI dataset reveals the effects of spatial and temporal correlations, as the BG and FG structures remain consistent across adjacent frames and scenes. ALDI++, having been trained with access to target domain information and learned BG itself features, leverages these correlations effectively. In contrast, ``ResM"—trained without such access—shows reduced performance on Virtual KITTI but regains superiority when BG perturbations (via BG-20K) disrupt these correlations. This highlights that models relying on spurious context may perform well under correlated conditions but generalize poorly in randomized settings.

Although mask pooling improves F1 scores in most cases, the gains are not uniform across all classes. For example, while "EffM" exhibited significant improvements across nearly all FG objects (76 of 77 pairs), some classes in ``ResM" (e.g., bicycle, motorcycle) saw marginal or even negative differences. This variation suggests that certain object classes may inherently depend more on contextual cues, or that architectural differences impact pooling effectiveness across object categories to make synergy.

Our findings indicate that ALDI++ learns strong FG-BG associations, likely due to its multi-stage training on foggy variants of Cityscapes. Its insufficient robustness under BG perturbation and high performance on Virtual KITTI  support this, reinforcing the idea that models trained with target domain context implicitly encode BG priors.


\subsection{Limitations}
Our approach has a significant limitation by being reliant on FG masks. Ablation study indicates it can cause significant performance drop by incorrect boundary. This is a classic ``chicken-or-egg" problem since we can generate masks from learnt features or we can utilize masks to separate FG and BG features as we have done in this work. There are a few approaches to generate the masks during training and inference - we could utilize an universal segmentation algorithm such as Segment Anything \cite{kirillov2023segment} to create such masks, utilize attention-based model \cite{huang2021causal,wang2021causal} or weak supervision \cite{wu2022background, zhai2024background, meng2021foreground}  or traditional boundary segmentation algorithms \cite{najman1994watershed} which do not utilize any learnt features to determine object boundaries. But none of the approaches are without errors and we aim to investigate this in our future work.

\bibliographystyle{plain}
\bibliography{references}
\end{document}